\pdfoutput=1

\documentclass[11pt]{article}

\usepackage[preprint]{acl}

\usepackage{times}
\usepackage{latexsym}

\usepackage[T1]{fontenc}

\usepackage[utf8]{inputenc}

\usepackage{microtype}

\usepackage{inconsolata}

\usepackage{graphicx}

\usepackage{booktabs}
\usepackage{array}
\usepackage{amsmath}
\usepackage{longtable}
\usepackage{enumitem}
\usepackage{multirow}
\usepackage{makecell}
\usepackage{graphicx}
\usepackage{subcaption}
\usepackage{epstopdf}
\usepackage{xcolor}
\usepackage{tcolorbox}
\usepackage{mdframed}
\usepackage{tabularx}
\usepackage{booktabs}
\usepackage{array}
\usepackage{soul}
\usepackage{geometry}
\usepackage{longtable}
\usepackage[table]{xcolor}
\tcbuselibrary{listings, breakable, skins}
\usepackage{adjustbox}

\title{Unsupervised Text Style Transfer for Controllable Intensity}

\author{Shuhuan Gu\textsuperscript{\rm 1},\ 
Wenbiao Tao\textsuperscript{\rm 1},\ 
Xinchen Ma\textsuperscript{\rm 1},\ 
Kangkang He\textsuperscript{\rm 2},\ 
Ye Guo\textsuperscript{\rm 2},\\ 
{\bf Xiang Li\textsuperscript{\rm 1},\ 
Yunshi Lan\textsuperscript{\rm 1}}\thanks{\ \  Corresponding author} \\ 
\textsuperscript{\rm 1}East China Normal University, Shanghai, China \\
\textsuperscript{\rm 2}Baowu Group, Shanghai, China\\
  \texttt{ treegu2003@gmail.com} \\
  \texttt{ yslan@dase.ecnu.edu.cn} \\
}

\begin{document}
\maketitle
\begin{abstract}
Unsupervised Text Style Transfer (UTST) aims to build a system to transfer the stylistic properties of a given text without parallel text pairs.
Compared with text transfer between style polarities, UTST for controllable intensity is more challenging due to the subtle differences in stylistic features across different intensity levels.
Faced with the challenges posed by the lack of parallel data and the indistinguishability between adjacent intensity levels, we propose a SFT-then-PPO paradigm to fine-tune an LLM.
We first fine-tune the LLM with synthesized parallel data.
Then, we further train the LLM with PPO, where the rewards are elaborately designed for distinguishing the stylistic intensity in hierarchical levels.
Both the global and local stylistic features are considered to formulate the reward functions.
The experiments on two UTST benchmarks showcase that both rewards have their advantages and applying them to LLM fine-tuning can effectively improve the performance of an LLM backbone based on various evaluation metrics.
Even for close levels of intensity, we can still observe the noticeable stylistic difference between the generated text.
\end{abstract}

\section{Introduction}

\begin{figure}[t]
\centering
\includegraphics[width=0.49\textwidth]{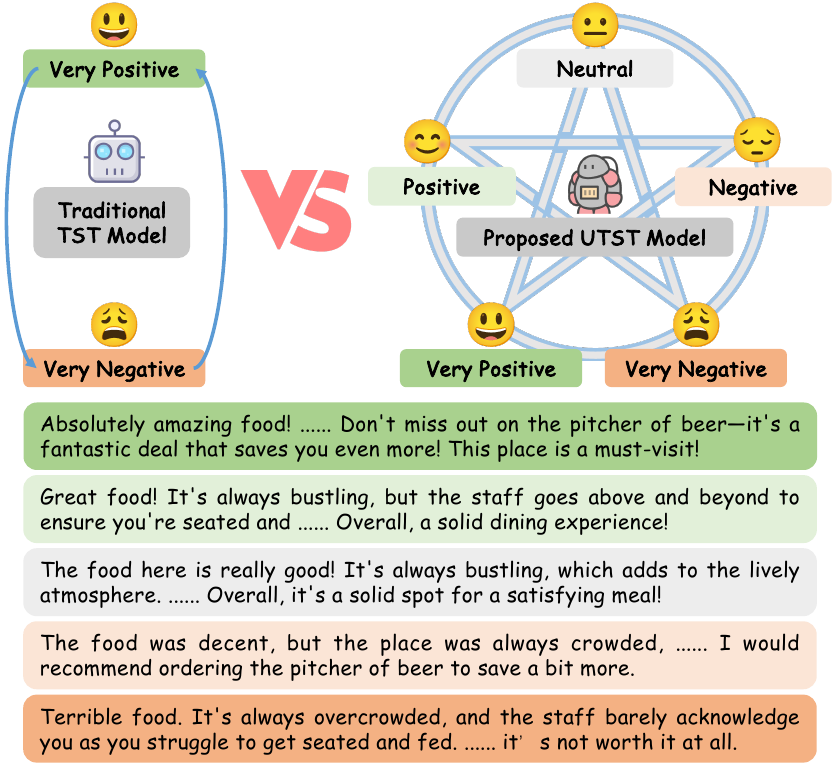}
\caption{A comparison between the text transfer for style polarity and across levels of intensity.}
\label{fig:motivation}
\end{figure}

Text Style Transfer (TST) aims to automatically modify the stylistic properties of a given text, such as readability~\cite{north2023lexical}, sentiment~\cite{DBLP:journals/corr/ZhangZL15}, or formality~\cite{briakou2021evaluating}, while preserving its original semantic content. 
This task, translating the original text into the desired style, has a wide range of applications in education~\cite{tan2024llm}, legal~\cite{li2021tst}, and social media~\cite{prabhumoye-etal-2018-style} fields. %
Besides the transfer between style polarity, recent trends start to evaluate the style transfer on a scale of intensity, which is more challenging due to the subtle difference of stylistic attributions and features for different levels of intensity~\cite{mir-etal-2019-evaluating,mukherjee-etal-2025-evaluating}.
Moreover, due to the lack of large high-quality parallel corpora, it brings in new difficulties in text style transfer without supervision, namely Unsupervised Text Style Transfer (UTST).
We display an example of UTST for controllable intensity in Figure~\ref{fig:motivation}.

Noticeable efforts have been made in existing studies. We classify them into two categories:
(1) \textit{Attribute substitution}. These methods replace the style-specific phrases in the text with the ones in the target style. 
Hence, extra corpora~\cite{li2018delete, sudhakar2019transforming}, classifier model~\cite{lee2020stable, fu2018style} and mapping model~\cite{ijcai2019p711} are introduced into the transfer model. However, due to the direct substitution of attributes, the transferred sentences may suffer from semantic incoherence.
(2) \textit{Sentence generation}. Thanks to the emergence of LLMs, text styles can be transferred token by token. Prompt engineering~\cite{reif2022recipe, suzgun2022prompt}, instruction fine-tuning~\cite{mukherjee2023leveraging, zong2024proswitch} and reinforcement learning (RL) ~\cite{deng2022rlprompt, liu2024step} are applied so that language models learn to generate target style text. Nevertheless, faced with the need for controllable intensity generation, these methods frequently suffer from style deficiencies.

The controllable intensity of text transformation forces a model to fully understand different levels of stylistic features and conduct style transfer. 
In summary, we need to focus on two challenges:

\begin{itemize}[leftmargin=*]
    \item \textbf{(C1) Absence of the high-quality TST parallel data.} Paired text style transfer data allows the model to learn the precise transfer pattern. While in the absence of high-quality parallel data, the knowledge source of the model is limited, which impairs the TST for controllable intensity.

    \item \textbf{(C2) Transfer across indistinguishable intensities.} Widely publicized bipolar style transformations have distinctive features for identification. However, there are more subtle differences between texts of stepped intensities, as shown in Figure~\ref{fig:motivation},
    which clearly demonstrates the UTST task. %
\end{itemize}

To address these challenges, we propose a SFT-then-PPO paradigm for UTST. Our approach begins by constructing a pseudo parallel corpus using an LLM to rewrite source texts into target styles at varying intensity levels. The generated outputs are then filtered using a pretrained style classifier to ensure alignment with the desired stylistic strength, resulting in high-quality pseudo-parallel corpora. Based on the synthesized data, we fine-tune an existing LLM to establish foundational TST capabilities. To further improve stylistic consistency and enable fine-grained control beyond supervised learning, we introduce proximal policy optimization with hierarchical rewards. Specifically, we design sentence-level rewards to capture global stylistic alignment, and lexicon-level rewards to guide fine-grained lexical choices. This dual-level reward structure allows the model to produce outputs with more accurate and consistent stylistic intensity.

To sum up, our contributions are as follows:

\begin{itemize}
    \item We introduce an intensity‑controlled UTST paradigm that first fine‑tunes a model for baseline style transfer via supervised fine-tuning and then refines it with reinforcement learning to further handle varying style intensities.
    \item We construct pseudo parallel corpus to solve the problem of scarcity of parallel data for TST model fine-tuning, as well as design sentence-level, lexicon-level and semantic-level rewards to allow the model to adapt to UTST for controllable intensity from multiple granularities.
    \item We construct two benchmarks for our task.
    The experimental results on readability and sentiment style transfer demonstrate that our method can effectively generate text with noticeable stylistic difference even for close levels of intensity.
\end{itemize}

\begin{figure*}[htbp]
\centering
\includegraphics[width=0.98\textwidth]{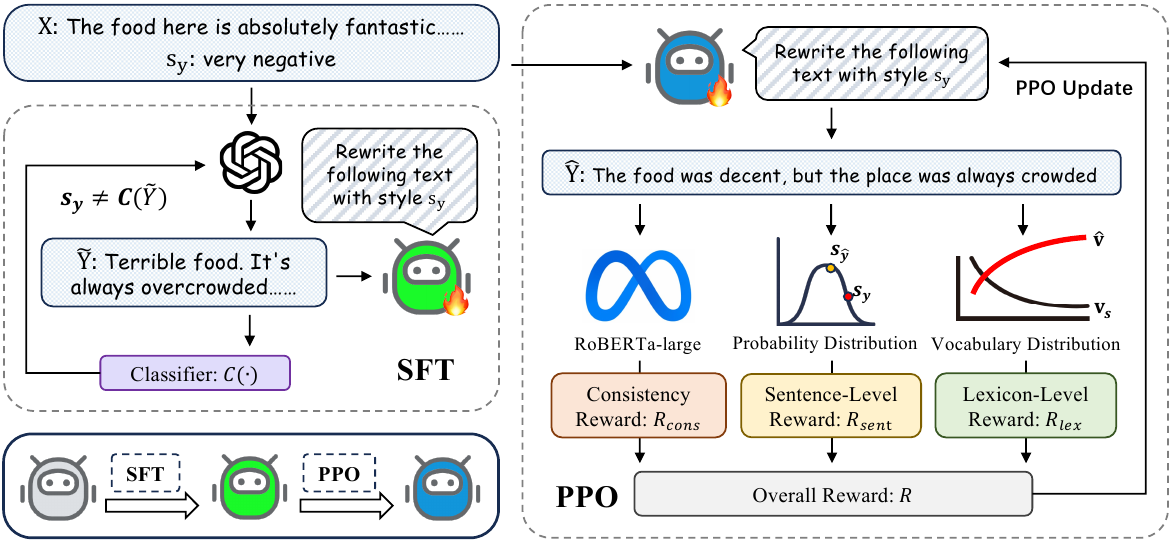}
\caption{UTST with SFT-then-PPO paradigm.}
\label{fig:method}
\end{figure*}

\section{Related Work}

\paragraph{Text Style Transfer.}
Text Style Transfer modifies text style while preserving meaning, typically via attribute substitution or sentence generation.

\textit{Attribute substitution} methods replace style-specific phrases using external corpora~\cite{li2018delete, sudhakar2019transforming}, classifiers~\cite{lee2020stable, fu2018style} or mapping models~\cite{ijcai2019p711}. While effective for coarse-grained polarity transfer, these methods often suffer from semantic incoherence and cannot handle fine-grained intensity---directly replacing phrases(e.g., `good' $\rightarrow$ `bad') fails to capture intensity differences(e.g., `great' vs. `amazing')~\cite{mir-etal-2019-evaluating}.

\textit{Sentence generation} methods based on LLMs are classified into non-disentangled and disentangled approaches. Non-disentangled methods (e.g., prompt engineering~\cite{reif2022recipe, suzgun2022prompt}, instruction fine-tuning~\cite{mukherjee2023leveraging, zong2024proswitch}) adopt intertwined style-content representations, which simplifies the training process yet lacks the capability to regulate subtle stylistic intensity. RL extensions~\cite{deng2022rlprompt, liu2024step} only optimize broad style alignment rather than nuanced gradations of style. Disentangled methods~\cite{han-etal-2024-disentangled} decouple style and content through auxiliary tasks to achieve precise control, albeit requiring extensive manual annotation.

Our paradigm follows the line of non-disentangled methods, enabling robust UTST by generating pseudo-parallel data with LLMs and applying hierarchical PPO-based RL for fine-grained style control, without needing parallel data.

\paragraph{Controllable Text Generation.}

Controllable text generation (CTG) produces text with desired attributes. 
Transformer-based LLMs have advanced CTG through prompt-based methods~\cite{suzgun2022prompt}, allowing flexible control via natural language instructions. Fine-tuning~\cite{mukherjee2023leveraging} and reinforcement learning (RL)~\cite{deng2022rlprompt, liu2024step} further improve adaptability for tasks like sentiment modification~\cite{DBLP:journals/corr/ZhangZL15} and readability adjustment~\cite{north2023lexical}. However, CTG for UTST with controllable intensity remains difficult due to the lack of parallel data and the challenge of capturing nuanced stylistic differences~\cite{mukherjee-etal-2025-evaluating}. 
Our approach fills these gaps by combining pseudo-parallel data generation with hierarchical rewards in a PPO method, enabling fine-grained style control and strong UTST performance without supervision.

\section{Task Statement}

UTST is an NLP task aiming to convert the style of the given text without parallel text pairs.
Specifically, given $k$ sets of non-parallel corpora, each of them can be represented as $\mathcal{D}^{s}= \{X_i\}_{i=1}^{|\mathcal{D}^{s}|}$.
Here, $s$ is annotated with a pre-defined stylistic intensity set $\mathcal{S} = \{s_1, s_2, .., s_k\}$, which is the categorical scale of the stylistic transfer.
Given a text $X = (x_1, x_2, ..., x_n)$ in the \textit{source} stylistic intensity $s_x \in \mathcal{S}$, the key of UTST is to develop a model which is able to transfer it into the text $Y =(y_1, y_2, ..., y_m)$ in the \textit{target} stylistic intensity $s_y \in \mathcal{S}$ while preserving the original content.

Regarding UTST for controllable intensity, the generation process can be modeled as:
\begin{align*}
    \hat{Y} \gets \mathcal{G}(X, s_y),
\end{align*}
where $\mathcal{G}(\cdot)$ is a UTST model and $\hat{Y}$ should be the transferred text in $s_y$ stylistic intensity.

\section{Paradigm for UTST}

To train such a model $\mathcal{G}(\cdot)$ without parallel corpus, we take full advantage of the LLMs for UTST with controllable intensity and introduce a SFT-then-PPO paradigm to train an LLM with surrogate rewards in hierarchical levels. 
Particularly, we first generate high-quality pseudo parallel corpus with an LLM-based pipeline.
Then, we fine-tune an LLM as $\mathcal{G}(\cdot)$ with parallel stylistic text.
Next, we further train the LLM with Proximal Policy Optimization (PPO), where the rewards are elaborately designed for distinguishing the fine-grained stylistic intensity.
The overview of our method is illustrated in Figure~\ref{fig:method}.
We will discuss the steps of obtaining $\mathcal{G}(\cdot)$ in detail.

\subsection{SFT with Synthesized Parallel Text}
\label{sec:sft}

SFT is critical for enhancing the text style transfer capabilities of LLMs. While LLMs exhibit impressive zero-shot and few-shot performance in TST tasks, their outputs often lack consistency in adhering to specific stylistic constraints, such as targeted readability levels or sentiment tones \cite{suzgun2022prompt, DBLP:journals/corr/abs-2010-12742}. SFT addresses this by accurately aligning the LLMs' outputs with the attributes of the target text, especially when the style is annotated with the fine-grained intensity.
To this end, we employ instruction fine-tuning~\cite{zhang2023instruction}, where LLMs are fine-tuned on elaborately designed prompts that specify the attributes of the target style.

Since SFT requires a high-quality dataset of parallel text pairs as the strong supervision, where each pair consists of an input text and its stylistically transformed counterpart. 
Due to the absence of the parallel text pairs for UTST task, we propose to synthetically generate the data for fine-tuning with the following pipeline.
\begin{itemize}[leftmargin=*]
    \item For each source text $X$ sampled from the corpus, we pair it with a variety of intensities in $\mathcal{S}$ to form a set of inputs $(X, s_y)$, where $s_y \in \mathcal{S}$ is the target style that we would like to transfer $X$ into and the transferred style $s_y \neq s_x$.
    \item Given an input, we utilize GPT-4o-mini~\cite{openai2024gpt4ocard} to transfer $X$ into the text with the target stylistic intensity $\tilde{Y}$\footnote{We display the prompts in Appendix~\ref{sec:data generation prompts}.}.
    \item We train a classifier $\mathcal{C}(\cdot)$ using the non-parallel corpus as the judge to determine if the synthesized target text satisfies the requested intensity.
    If the synthesized text $\tilde{Y}$ is predicted as the target intensity, we add the parallel data $(X, s_y, \tilde{Y})$ into the data set.
    Otherwise, we repeat the above step till the LLMs generate the desired text. 
    In the case where no suitable output is generated after $10$ iterations, we discard $X$.
\end{itemize}

As a result, we obtain a set of parallel data $\mathcal{D} = \{X_i, s_y, \tilde{Y}_i\}_{i=1}^N$ which contains aligned text pairs for style transfer. 
We consider an LLM as $\mathcal{G}(\cdot)$ and fine-tune it on $\mathcal{D}$ with the following objective:
\begin{align*}
    \mathcal{L}_{\text{SFT}} = -\sum_{i = 1}^{N} \sum_{j = 1}^{k}\log{P(\tilde{Y}_i | X_i, s_j)},
\end{align*}
where the probability is modeled via the LLM.
After fine-tuning, the model $\mathcal{G}(\cdot)$ is able to generate text $\hat{Y}$ in the desired style.

\subsection{Reward Function for PPO}
\label{sec:ppo}

While SFT effectively guides the model to capture stylistic features through strong supervision, it often falls short in ensuring consistent adherence to subtle or nuanced style attributes across diverse inputs. To address this, we introduce an RL-based approach to refine the model’s text generation, inspired by prior RL methods \cite{gong-etal-2019-reinforcement, Luo19DualRL}. Unlike prior RL-based methods that use single-scale rewards to judge polarity compliance (e.g., positive vs. negative), a contribution of our method lies in its hierarchical reward optimization quantifying intensity compliance (e.g., very positive vs. positive), which explicitly evaluates the global and local attributes of the stylistic text.
We introduce two-level rewards: a sentence-level reward that measures the stylistic intensity of the generated text in two different formats, and a lexicon-level reward that aligns the generated text to the fine-grained linguistic features by analyzing lexical distributions.
We integrate the above rewards into our final reward function.

\subsubsection{Hierarchical Stylistic Rewards}
\label{hierarchical rewards}

\noindent \textbf{Sentence-level Reward}. 
The sentence-level reward, derived from classifier-based evaluations, captures global stylistic attributes.
Since the stylistic intensity can be either formulated as the fine-scale classification (e.g., sentiment analysis~\cite{DBLP:journals/corr/ZhangZL15}) or regression (e.g., readability assessment~\cite{feng2010comparison}),
we formulate the sentence-level rewards for both regression and classification.

\begin{itemize}[leftmargin=*]
    \item \emph{Regression-based}: This type of reward is designed for the case that a regression model $\mathcal{C}(\cdot)$ is used to measure the intensity of the generated text $\hat{Y}$.
    A large value reward is provided when the generated text has the low deviation from the target intensity.
    Otherwise, the reward value is small.
    Inspired by existing study~\cite{ribeiro-etal-2023-generating}, we model the reward for regression-based classifier using a normalized Gaussian distribution centered at the target intensity $f \sim \mathcal{N}(s_y, \sigma^2)$:
    \begin{align*}
        \hat{s} & = \mathcal{C}(\hat{Y}); \\
        \mathcal{R}_{\text{sent}}(\hat{Y}) & = f(\hat{s})/f(s_y).
    \end{align*}
    The reward will assign a maximum value $1$ if the observed intensity is equal to the target intensity and decrease exponentially as $\hat{s}$ deviates from $s_y$.
    \item \emph{Classification-based}: This type of reward is designed for the case that a classification model $\mathcal{C}(\cdot)$ is used to measure the intensity of the generated text.
    The reward is formulated as:
    \begin{align*}
        \mathcal{R}_{\text{sent}}(\hat{Y})  = \mathcal{C}(\hat{Y}),
    \end{align*}
    where $\mathcal{C}(\hat{Y})$ outputs the probability of the target intensity given the generated text.
    The reward will assign a maximum value $1$ to the generated text if it is predicted to have the target style.
\end{itemize}

As we can see, both rewards encourage the model to generate text in the target stylistic intensity.

\noindent \textbf{Lexicon-level Reward}.
The sentence-level reward focuses on overall characteristics while lexicon-level reward captures vocabulary intensity from a style-specific vocabulary and lexical distribution.
Since stylistic distinctions often manifest through different word usage patterns~\cite{mir-etal-2019-evaluating}, identifying and rewarding style-specific vocabulary can enhance the LLMs' ability to generate text that conforms to the linguistic features of the target style.

To identify the style-specific vocabulary, we employ TF-IDF to measure the importance of a word to a text in a collection or corpus.
As a result, for the UTST task, the style-specific words are usually those having high TF-IDF values.
We select the words with top-$1000$ TF-IDF values per level of intensity to form the style-specific vocabulary.
We display the representative words in a corpus for stylistic intensities in Appendix~\ref{sec:representative words}.
Then for each level of stylistic intensity, we concatenate all the text in the corpus and form a TF-IDF vector demonstrating a pivot of the lexical distribution.
For each generated text, we also form a TF-IDF vector showing its lexical distribution:
\begin{align*}
    \mathbf{v}_s &= \text{TF-IDF}(\mathcal{D}^s); \\
    \hat{\mathbf{v}} &= \text{TF-IDF}(\hat{Y}),
\end{align*}
where $\mathbf{v}_s$ indicates the pivot TF-IDF vector for the requested intensity $s$ and $\hat{\mathbf{v}}$ means the TF-IDF vector for the generated text.
By computing the similarity of the vectors of the generated text and the pivot vector, the lexicon-level reward can be measured as follows:
\begin{align*}
   \mathcal{R}_{\text{lex}} = \frac{\exp (\text{sim}(\mathbf{v}, \mathbf{v}_{s_y})/T)}{\sum_{j=1}^{k} \exp(\text{sim}(\hat{\mathbf{v}}, \mathbf{v}_{s_j}) / T)},
\end{align*}
where $\mathbf{v}_{s_y}$ is the pivot TF-IDF vector of the target intensity, $\text{sim}(\cdot, \cdot)$ denotes the cosine similarity function, and $T$ is a temperature parameter that controls the sharpness of the distribution.

This reward encourages the model to emphasize vocabulary patterns that are characteristic of the target style, thereby improving stylistic fidelity.

\subsubsection{Consistency Reward}

To preserve the semantic information of the input during style transfer, we follow the existing study~\cite{bert-score} proposing a consistency reward:

\begin{align*}
    \mathcal{R}_{\text{cons}} = \frac{1}{|\hat{Y}|} \sum_{y_j \in \hat{Y}} \max_{x_i \in X} (\text{Emb}(y_j) \cdot \text{Emb}(x_i)),
    \label{reward-bertscore}
\end{align*}
where $|\hat{Y}|$ is the number of tokens in $\hat{Y}$, $\text{Emb}(\cdot)$ denotes the RoBERTa~\cite{DBLP:journals/corr/abs-1907-11692} embeddings of token. This score is normalized to \([0, 1]\), which indicates the semantic alignment between the source and target text.

\subsubsection{Total Reward}
\label{sec:total reward}

The total reward is formulated as the weighted sum of the above rewards:
\begin{align*}
    \mathcal{R} = \lambda_1 \cdot \mathcal{R}_{\text{sent}} + \lambda_2 \cdot \mathcal{R}_{\text{lex}} + \lambda_3 \cdot \mathcal{R}_{\text{cons}},
\end{align*}
where $\lambda_1$, $\lambda_2$ and $\lambda_3$ are three hyper-parameters to balance the rewards.
Our final model is obtained via the fine-tuning in Section~\ref{sec:sft} and PPO optimization.
As a result, our model $\mathcal{G}(\cdot)$ can perform text style transfer for controllable intensity without any parallel data.

\section{Experimental Setup}

\subsection{Datasets}

Differently from most previous work, which only focused on the text style transfer between style polarity (e.g., positive $\rightarrow$ negative), we conduct experiments on two datasets\footnote{The details of the datasets are shown in the Appendix~\ref{ap:dataset}} of UTST for controllable intensity:

\begin{itemize}[noitemsep,topsep=0pt,leftmargin=6mm]
    \item \textbf{\textsc{Cnn/dm}}: This is a dataset for readability transfer which is collected from CNN Daily mails \cite{ DBLP:conf/nips/HermannKGEKSB15} containing the original text and its summaries with varying readability levels. 
    To construct a dataset for UTST task, we calculate the Flesch Reading Ease (FRE) scores~\cite{flesch1948new} of the summaries and divide them into four levels of readability, namely \{\textit{elementary school (1)} $\rightarrow$ \textit{middle school} (2) $\rightarrow$ \textit{high school} (3) $\rightarrow$ \textit{college} (4)\}.
    The numbers of text in these sets are $42824$, $176262$, $61099$ and $6459$, respectively.
    \item \textbf{\textsc{Yelp}}: This is a dataset initially for sentiment analysis which is collected from Yelp reviews \cite{DBLP:journals/corr/ZhangZL15}.
    It contains the ratings ranging from $1$ to $5$ stars, which corresponds to \{\textit{very negative (1)} $\rightarrow$ \textit{negative} (2) $\rightarrow$ \textit{neutral} (3) $\rightarrow$ \textit{positive} (4) $\rightarrow$ \textit{very positive} (5)\}. 
    We consider it as the UTST dataset and partition it into five sets of non-parallel corpus for training.
    The number of texts in each set is $130$k.
\end{itemize}

\subsection{Evaluation Metrics}

Following the standard evaluation of TST tasks, we employ metrics that assess style alignment and content preservation. 
Regarding style alignment, we use \textbf{FRE} to measure the readability intensity of the prediction and \textbf{FRE}$_\Delta$ to denote the deviation of intensity between the prediction and the target.
We use star rating (\textbf{STAR}) to measure the sentiment intensity of the prediction and \textbf{STAR}$_\Delta$ to denote the deviation of intensity between the prediction and the target. 
We also employ the sum of the hierarchical reward \textbf{H-Re}, illustrated in Section~\ref{sec:ppo} to measure the stylistic alignment of the test data.
For content preservation, ROUGE-L (\textbf{RG-L}) is used to measure the overlap between the prediction and the input text. 
Detailed description can be found in Appendix~\ref{Evaluation Metrics}.

\subsection{Comparable Methods}

To evaluate the performance of our proposed UTST method with controllable intensity, we comprehensively compare a variety of our SFT-then-PPO paradigms for comparison:
\textbf{SFT}: We employ a flan-t5-large model as the backbone LLM and fine-tune it via the SFT with the synthesized parallel text. 
\textbf{SFT + PPO ($\mathcal{R}_{\text{sent}} + \mathcal{R}_{\text{cons}}$)}: This is our method with SFT-then-PPO paradigm and we employ sentence-level reward as the style alignment judge.
\textbf{SFT + PPO ($\mathcal{R}_{\text{lex}} + \mathcal{R}_{\text{cons}}$)}: This is our method with SFT-then-PPO paradigm and we employ lexicon-level reward as the style alignment judge.
\textbf{SFT + PPO ($\mathcal{R}_{\text{sent}} + \mathcal{R}_{\text{lex}} + \mathcal{R}_{\text{cons}}$)}: This is our method with both hierarchical rewards as the style alignment judge.

Due to the inapplicability of the prior TST methods for polarity transfer, we employ \textbf{GPT-4o-mini}~\cite{openai2024gpt4ocard}, which is an advanced LLM, as the strong baseline for UTST tasks.
We simply prompt it with instructions to generate text with varying style intensities under the zero-shot setting, as detailed in Appendix~\ref{sec:evalution prompts}.

\subsection{Implementation Details}

For \textsc{Yelp}, we employ a DistilBERT\footnote{\url{https://huggingface.co/distilbert}} fine-tuned on non-parallel corpus as the classification-based model $\mathcal{C}(\cdot)$.
For \textsc{Cnn/dm}, we employ FRE function, which is a regression model, as $\mathcal{C}(\cdot)$.

We adopt the TRLX library \cite{havrilla-etal-2023-trlx} with PPO for the RL phase. Reward weights are set to $\lambda_1 = 0.5$, $\lambda_2 = 0.3$, and $\lambda_3 = 0.2$ for readability and sentiment, explained in Appendix~\ref{setting_strategy}, with a softmax temperature $T = 0.01$ for the lexicon-level reward. 
Detailed parameter settings can be seen in Table~\ref{tab:hyperparameters}.

\section{Results}
\subsection{Main Results}

\noindent \textbf{Readability Transfer}.
Table~\ref{readability_results} compares the UTST paradigm with baselines on the \textsc{Cnn/dm} dataset across four target readability levels. Without task-specific training, GPT-4o-mini underperforms in zero-shot settings, highlighting the limitations of prompt-based approaches—even with a strong LLM. In contrast, our UTST paradigm achieves significant gains, outperforming baselines by up to $64.05\%$ in average $\text{FRE}_{\Delta}$ and $40.74\%$ in H-Re, demonstrating the effectiveness of the SFT-then-PPO pipeline.

\begin{table}[h]
  \small
  \centering
  \setlength{\tabcolsep}{1.5pt}
  \begin{tabular}{lcccc}
    \toprule
    \textbf{Readability ($s_y$)} & \textbf{FRE} & \textbf{FRE}$_\Delta$ $\downarrow$ & \textbf{RG-L} $\uparrow$ & \textbf{H-Re} $\uparrow$\\
    \midrule
    \multicolumn{5}{c}{\textbf{GPT-4o-mini}} \\
    \midrule
    Elementary school & 70.96 & 19.04 & 54.12 & 0.24\\
    Middle school & 65.41 & 4.59 & 59.64 & 0.40\\
    High school & 60.73 & 10.73 & 62.54 & 0.30\\
    College & 47.42 & 27.42 & 60.92 & 0.14 \\
    Average & - & 15.44 & \underline{59.31} & 0.27\\
    \midrule
    \multicolumn{5}{c}{\textbf{SFT}} \\
    \midrule
    Elementary school & 73.97 & 16.03 & 61.63 & 0.29\\
    Middle school & 64.96 & 5.04 & 63.38 & 0.40\\
    High school & 51.89 & 1.89 & 61.91 & 0.38\\
    College & 32.27 & 12.27 & 55.22 & 0.29\\
    Average & - & 8.81 & \textbf{60.54} & 0.34\\
    \midrule
    \multicolumn{5}{c}{\textbf{SFT + PPO ($\mathcal{R}_{\text{sent}} + \mathcal{R}_{\text{cons}}$)}} \\
    \midrule
    Elementary school & 78.77 & 11.23 & 58.62 & 0.38\\
    Middle school & 65.25 & 4.75 & 63.11 & 0.40\\
    High school & 50.70 & 0.70 & 61.27 & 0.38\\
    College & 28.87 & 8.87 & 53.06 & 0.32\\
    Average & - & 6.39 & 59.02 & \underline{0.37}\\
    \midrule
    \multicolumn{5}{c}{\textbf{SFT + PPO ($\mathcal{R}_{\text{lex}} + \mathcal{R}_{\text{cons}}$)}} \\
    \midrule
    Elementary school & 79.96 & 10.04 & 56.73 & 0.40\\
    Middle school & 64.65 & 5.35 & 61.65 & 0.40\\
    High school & 48.92 & 1.08 & 59.50 & 0.38\\
    College & 25.71 & 5.71 & 49.99 & 0.34\\
    Average & - & \textbf{5.55} & 56.97 & \textbf{0.38}\\
    \midrule
    \multicolumn{5}{c}{\textbf{SFT + PPO ($\mathcal{R}_{\text{sent}} + \mathcal{R}_{\text{lex}} + \mathcal{R}_{\text{cons}}$)}} \\
    \midrule
    Elementary school & 78.87 & 11.13 & 58.12 & 0.38\\
    Middle school & 65.25 & 4.75 & 62.69  & 0.40\\
    High school & 50.89 & 0.89 & 60.90 & 0.38\\
    College & 26.67 & 6.67 & 51.49 & 0.33\\
    Average & - & \underline{5.86} & 58.30 & \underline{0.37}\\
    \midrule
  \end{tabular}
  \caption{The results of \textsc{cnn/dm}. Both breakdown results and average results are displayed.}
  \label{readability_results}
\end{table}

\begin{table}[t]
  \small
  \centering
  \setlength{\tabcolsep}{3pt}
  \begin{tabular}{lcccc}
    \toprule
    \textbf{Sentiment ($s_y$)} & \textbf{STAR} & \textbf{STAR}$_\Delta$ $\downarrow$ & \textbf{RG-L} $\uparrow$ & \textbf{H-Re} $\uparrow$\\
    \midrule
    \multicolumn{5}{c}{\textbf{GPT-4o-mini}} \\
    \midrule
    Very Negative & 1.5190 & 0.5190 & 45.39 & 0.35 \\
    Negative & 1.8018 & 0.1982 & 47.11 & 0.38\\
    Neutral & 2.9476 & 0.0524 & 46.58 & 0.23\\
    Positive & 4.3652 & 0.3652 & 49.78 & 0.30\\
    Very Positive & 4.6032 & 0.3968 & 48.98 & 0.41\\
    Average & - & 0.3063 & 47.57 & 0.33\\
    \midrule
    \multicolumn{5}{c}{\textbf{SFT}} \\
    \midrule
    Very Negative & 1.2012 & 0.2012 & 43.68 & 0.50\\
    Negative & 2.3338 & 0.3338 & 45.88 & 0.38\\
    Neutral & 3.3854 & 0.3854 & 46.24 & 0.37\\
    Positive & 4.1382 & 0.1382 & 46.20 & 0.32\\
    Very Positive & 4.9416 & 0.0584 & 43.92 & 0.60\\
    Average & - & 0.2234 & 45.18 & 0.43\\
    \midrule
    \multicolumn{5}{c}{\textbf{SFT + PPO ($\mathcal{R}_{\text{sent}} + \mathcal{R}_{\text{cons}}$)}} \\
    \midrule
    Very Negative & 1.1954 & 0.1954 & 46.16 & 0.50\\
    Negative & 2.3254 & 0.3254 & 48.55 & 0.38\\
    Neutral & 3.3760 & 0.3760 & 48.95 & 0.38\\
    Positive & 4.1330 & 0.1330 & 49.21 & 0.32\\
    Very Positive & 4.9374 & 0.0626 & 47.88 & 0.60\\
    Average & - & $\mathbf{0.2185}$ & $\mathbf{48.15}$ &  $\underline{0.44}$\\
    \midrule
    \multicolumn{5}{c}{\textbf{SFT + PPO ($\mathcal{R}_{\text{lex}} + \mathcal{R}_{\text{cons}}$)}} \\
    \midrule
    Very Negative & 1.2004 & 0.2004 & 46.19 & 0.50\\
    Negative & 2.3146 & 0.3146 & 48.80 & 0.38\\
    Neutral & 3.3904 & 0.3904 & 49.14 & 0.38\\
    Positive & 4.1370 & 0.1370 & 49.21 & 0.32\\
    Very Positive & 4.9410 & 0.0590 & 46.53 & 0.60\\
    Average & - & $0.2203$ & $\underline{47.97}$ & \underline{0.44}\\
    \midrule
    \multicolumn{5}{c}{\textbf{SFT + PPO ($\mathcal{R}_{\text{sent}} + \mathcal{R}_{\text{lex}} + \mathcal{R}_{\text{cons}}$)}} \\
    \midrule
    Very Negative & 1.2114 & 0.2114 & 46.18 & 0.51\\
    Negative & 2.3234 & 0.3234 & 48.49 & 0.40\\
    Neutral & 3.3748 & 0.3748 & 48.89 & 0.39\\
    Positive & 4.1462 & 0.1462 & 49.04 & 0.33\\
    Very Positive & 4.9374 & 0.0374 & 46.39 & 0.62\\
    Average & - & $\underline{0.2186}$ & 47.80 & $\mathbf{0.45}$\\
    \bottomrule
  \end{tabular}
  \caption{The results of \textsc{yelp}. Both breakdown results and average results are displayed.
  }
  \label{sentiment_results}
\end{table}

\noindent \textbf{Sentiment Transfer}.
We further evaluate the UTST method on the \textsc{Yelp} dataset for sentiment transfer, with results presented in Table~\ref{sentiment_results}. Consistent with the readability transfer task, UTST significantly outperforms GPT-4o-mini, achieving gains of $26.85\%$ in $\text{STAR}_{\Delta}$ and $36.36\%$ in H-Re, reinforcing its effectiveness in text style transfer.

\noindent \textbf{Overall Analysis}. \textbf{H-Re} serves as the primary metric for UTST tasks, as it comprehensively assesses performance at both the sentence and lexical levels.
Overall, the SFT-then-PPO paradigm outperforms SFT alone (H-Re: 0.38 vs. 0.34 in Table~\ref{readability_results}; 0.45 vs. 0.43 in Table~\ref{sentiment_results}). SFT is only capable of learning to mimic the superficial stylistic characteristics of the dataset, and fails to proactively align with complex or continuous style control objectives. In contrast, PPO facilitates the direct optimization of non-differentiable style metrics via reward signals, thus improving control over fine-grained styles.

The optimal reward combination exhibits variation across different style types: for readability, $\mathcal{R}_{\text{lex}} + \mathcal{R}_{\text{cons}}$ is deemed optimal, as readability relies on lexical selections (captured by $\mathcal{R}_{\text{lex}}$) and content preservation (captured by $\mathcal{R}_{\text{cons}}$)---sentential features are redundant here, given that word difficulty exhibits a strong correlation with sentence structure. For sentiment, $\mathcal{R}_{\text{sent}} + \mathcal{R}_{\text{lex}} + \mathcal{R}_{\text{cons}}$ is considered optimal, as emotional meaning is context-dependent (e.g. 'busy' can convey positive or negative connotations), necessitating $\mathcal{R}_{\text{sent}}$ to ascertain the emotional polarity of words within their contextual framework. This adaptability highlights the generalizability of the proposed model across diverse style dimensions.

However, \textbf{RG-L} exhibits a slight reduction, attributed to the intrinsic calculation mechanism. A detailed explication is provided in Appendix~\ref{app:Rouge-L}.

\begin{figure*}[!t]
\centering
\begin{subfigure}[t]{0.32\textwidth}
    \centering
    \includegraphics[width=0.98\textwidth]{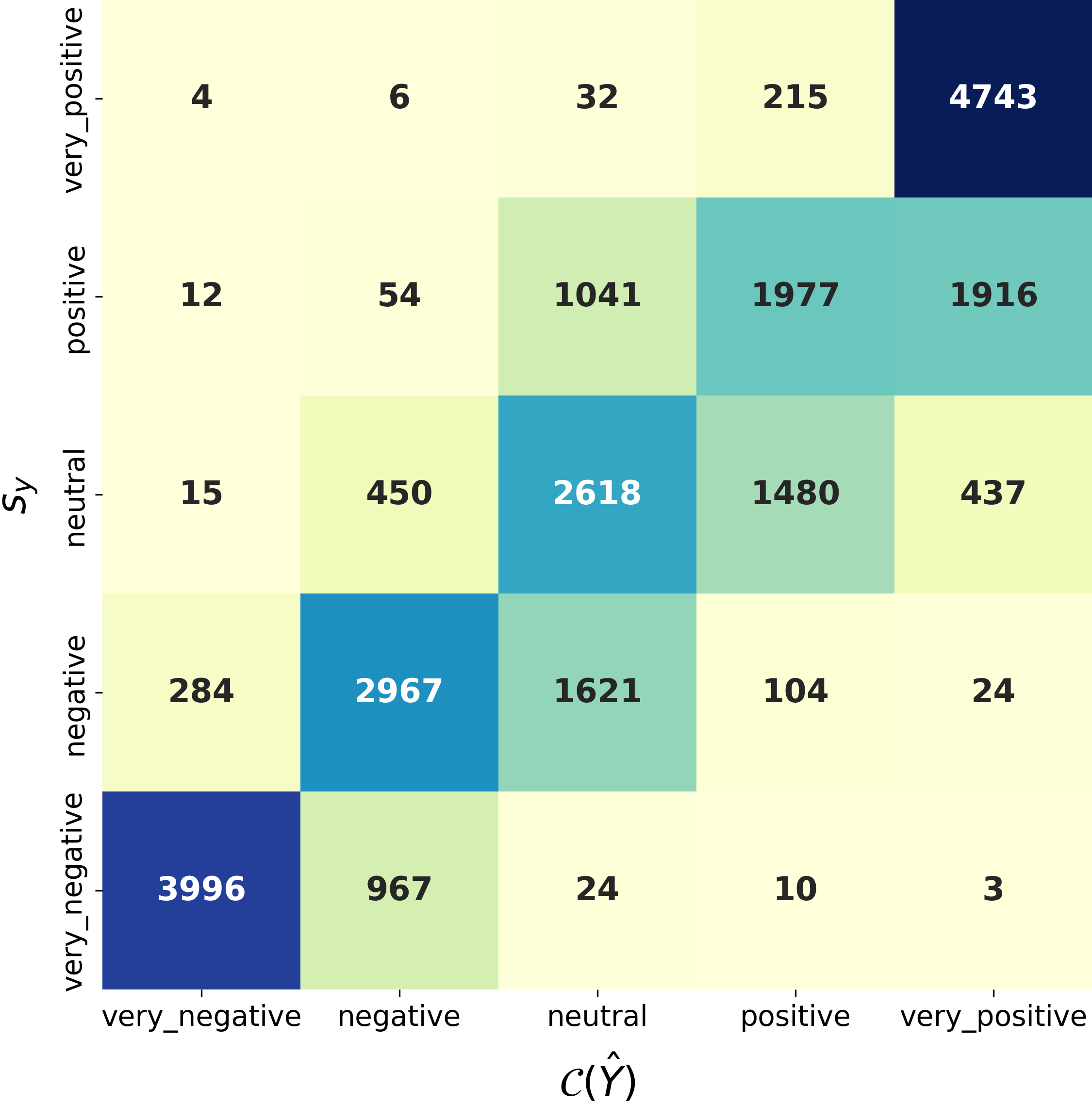}
    \caption{Heatmap of factual consistency on the \textsc{Yelp} dataset. Each cell indicates the number of instances. The sentiment of the transferred text is predicted via a fine-tuned DistilBERT classifier.}
    \label{sentiment_heatmap}
\end{subfigure}
\hspace{0.3em}
\begin{subfigure}[t]{0.62\textwidth}
    \centering
    \includegraphics[width=0.98\textwidth]{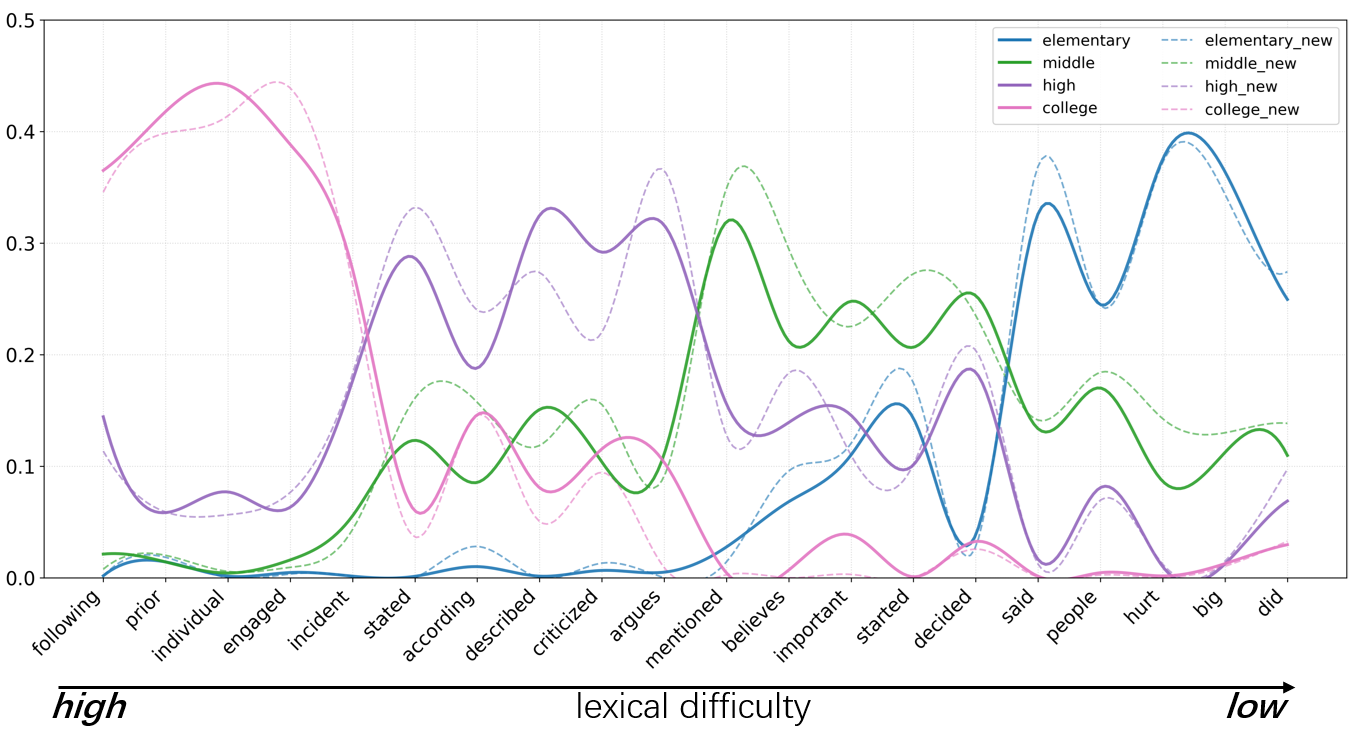}
    \caption{Visualization of the lexicon distribution. We showcase some style-specific words and their TF-IDF values in the pivot vectors and the generated text.
    The solid line denotes the distribution of the annotated corpus and the imaginary line denotes the distribution of the generated text.
    The words from left to right sides showcase a decreasing lexical difficulty.}
    \label{TF_IDF}
\end{subfigure}
\caption{Factual consistency and lexicon distribution analysis of SFT + PPO ($\mathcal{R}_{\text{sent}} + \mathcal{R}_{\text{lex}} + \mathcal{R}_{\text{cons}}$).}
\label{fig:combined_analysis}
\end{figure*}

\begin{table*}[!t]
\centering
\small
\begin{adjustbox}{max width=\textwidth}
\begin{tabular}{@{}l p{15cm}@{}}
\toprule
\textbf{Category} & \textbf{Text} \\
\midrule

\textbf{Original} & The food here is absolutely fantastic. It's always busy, which speaks to its popularity, but the staff are efficient and make sure you're seated and served quickly. Be sure to order the pitcher of beer—it's a great deal and will save you a good amount. \\
\hline

\textbf{Elementary} & The food here is really \colorbox{blue!10}{good}. It's always busy, which is \colorbox{blue!10}{good} because many \colorbox{blue!10}{people} come here. The staff is very \colorbox{blue!10}{helpful}. They make sure you get seated \colorbox{blue!20}{quickly} and \colorbox{blue!10}{get} your food. Don't forget to order a pitcher of beer. It's a \colorbox{blue!20}{great} deal and will save you a lot. \\

\textbf{Middle} & The food at this \colorbox{blue!20}{restaurant} is always \colorbox{blue!20}{great}, and it's easy to see why it's so \colorbox{blue!20}{popular}. The staff is always \colorbox{blue!20}{friendly} and ready to help, making sure your meal comes out \colorbox{blue!20}{quickly}. It's also a \colorbox{blue!10}{good} idea to order a pitcher of beer—it's a \colorbox{blue!20}{great} deal and can save you a lot of money. \\

\textbf{High} & The food at this \colorbox{blue!30}{establishment} is \colorbox{blue!30}{consistently} \colorbox{blue!30}{excellent}, and it's clear why it's so \colorbox{blue!20}{popular}. The service is \colorbox{blue!30}{prompt} and \colorbox{blue!30}{attentive}, ensuring that your meal is served  \colorbox{blue!30}{promptly}. \colorbox{blue!30}{Additionally}, it's a \colorbox{blue!10}{good} idea to order a pitcher of beer—it's a \colorbox{blue!20}{great} deal and can save you quite a bit of money. \\

\textbf{College} & The dining experience at this establishment \colorbox{blue!40}{epitomizes} a \colorbox{blue!40}{remarkable} level of quality and service. Despite its \colorbox{blue!40}{perpetually} \colorbox{blue!40}{bustling} \colorbox{blue!40}{atmosphere}, which reflects the \colorbox{blue!20}{restaurant's} \colorbox{blue!40}{considerable} appeal, the staff exhibits a remarkable level of \colorbox{blue!40}{efficiency}, ensuring \colorbox{blue!30}{prompt} service and \colorbox{blue!30}{prompt} attention to \colorbox{blue!40}{patrons}. \colorbox{blue!40}{Moreover}, it is \colorbox{blue!40}{imperative} to reserve a pitcher of lager—it offers an \colorbox{blue!40}{exceptional} value and \colorbox{blue!40}{significantly} reduces your overall \colorbox{blue!40}{expenditure}. \\

\bottomrule
\end{tabular}
\end{adjustbox}
\caption{Transferred text of a restaurant review with different levels of readability intensities. Highlighted words are those in style-specific vocabulary and the darker color indicates a high level of intensity.}
\label{tab: case}
\end{table*}

\subsection{Further Analysis}

\subsubsection{Factual Consistency Analysis}

The heatmap in Figure~\ref{sentiment_heatmap} reveals strong diagonal alignment in predictions, particularly for Very Positive (4743 cases) and Very Negative (3996 cases), demonstrating the model’s accuracy in identifying emotionally charged expressions. Most misclassifications occur between adjacent categories. For example, Positive as Very Positive (1916 cases), underscoring the challenge of distinguishing subtle emotional shifts. Neutral samples exhibit high ambiguity, often confused with neighboring categories (4985 instances total), likely due to overlapping lexical patterns. Despite these challenges, the model shows strong performance, effectively capturing both clear and nuanced emotional signals.

\subsubsection{Visualization of the Lexicon Distribution}

Figure~\ref{TF_IDF} compares lexicon distribution across readability intensity levels regarding both reference and generated texts. Results show high lexical consistency, especially at style polarity---vocabulary in generated texts closely mirrors that of references. Even at intermediate levels, differences remain minimal, highlighting the model’s ability to reproduce readability-specific vocabulary distributions. This strong lexical alignment, supported by the hierarchical reward design, further confirms the effectiveness of our paradigm in capturing fine-grained stylistic attributes.

\subsubsection{Case Study}

As shown in Table~\ref{tab: case}, we transform a restaurant review in \textsc{yelp} dataset into four versions with different readability levels. The text length increases progressively with higher reading difficulty. Furthermore, the number of highlighted words indicates that lower-readability versions tend to include more advanced vocabulary, such as ``\textit{epitomizes}'' and ``\textit{imperative}''. This example demonstrates the effectiveness and sensitivity of our method in generating text with varying stylistic intensities.

\section{Conclusion}
We propose an SFT-then-PPO paradigm for UTST with controllable intensity, using filtered pseudo-parallel corpora for SFT and hierarchical PPO for fine-grained style control. Experiments on CNN/DM and Yelp show improved readability and sentiment transfer, with accurate style alignment, strong content preservation, and enhanced lexical precision over GPT-4o-mini. This approach advances UTST and enables broader applications in text transformation. Future work will explore extending intensity control to additional stylistic dimensions for greater real-world applicability.

\section*{Limitations}

Our UTST paradigm exhibits promising performance in controllable intensity text style transfer, one potential limitation concerns the linguistic scope of the evaluation. The current experiments are conducted on English-language datasets (CNN/DM and Yelp), which, while representative of common style transfer benchmarks, may not fully capture the method's effectiveness across diverse linguistic and cultural contexts. Given that the proposed approach is inherently language-agnostic, future work could benefit from extending the evaluation to multilingual or non-English datasets, thereby providing a more comprehensive assessment of the model’s generalizability and practical utility in broader real-world scenarios.

\bibliography{anthology,custom}

\clearpage
\onecolumn
\appendix

\section{Prompts for Parallel Data Generation}
\label{sec:data generation prompts}

We present prompts for generating readability and sentiment parallel data.

\subsection{readability}
\begin{mdframed}
Rewrite the following text to match the $\{\text{outputDesc}\}$ readability level
(Flesch Reading Ease score $\{\text{FREScore}\}$).

Adjust vocabulary, sentence length, and complexity to strongly reflect the target readability level. 

For Elementary, use very simple words and short sentences.

For Middle School, use moderately simple words and slightly longer sentences.

For High School, use more complex words and varied sentence structures.

For College, use advanced vocabulary and complex sentence structures.

Keep the core meaning and content consistent, but adapt the style to be natural and concise. 

Input: $\{\text{text}\}$

Output:
\end{mdframed}

\subsection{sentiment}
\begin{mdframed}
Rewrite the following $\{\text{inputRating}\}$ Yelp review into an $\{\text{outputRating}\}$ review.  

Transform the tone and wording to reflect the target rating’s sentiment strongly.  

For 5 Stars, use highly positive words like "amazing" or "outstanding";  

for 1 Star, use strongly negative words like "terrible" or "awful".  

Keep core aspects (food, service, atmosphere) consistent, but adjust their descriptions to match the target rating.  

Ensure the output is natural, realistic, concise (similar length to the input), and avoids vague terms like "okay" or "fine".

Input: $\{\text{text}\}$

Output:
\end{mdframed}

\section{Prompts for Controllable Text Generation}
\label{sec:evalution prompts}

We present prompts for generating readability and sentiment text with controllable intensity.

\subsection{readability}
\begin{mdframed}
Rewrite the following text for $\{\text{readability level}\}$

Input: $\{\text{text}\}$

Output:
\end{mdframed}

\subsection{sentiment}
\begin{mdframed}

Rewrite the following text for $\{\text{sentiment level}\}$

Input: $\{\text{text}\}$

Output:
\end{mdframed}

\section{Evaluation Metrics}
\label{Evaluation Metrics}
\begin{itemize}[noitemsep,topsep=0pt,leftmargin=6mm]
    \item \textbf{FRE}: Measures readability using the Flesch Reading Ease score, assessing text complexity for readability transfer:
    \begin{align*}
        \text{FRE} = 206.835 - 1.015 \cdot \frac{\text{total words}}{\text{total sentences}} - 84.6 \cdot \frac{\text{total syllables}}{\text{total words}}.
    \end{align*}
    \textbf{FRE}$_\Delta$ quantifies deviation from the target range’s midpoint:
    \begin{align*}
        \text{FRE}_\Delta = |\text{FRE}_{\hat{y}} - \text{FRE}_{s_y}|,
    \end{align*}
    where \(\text{FRE}_{\hat{y}}\) is the FRE of the generated text, and \(\text{FRE}_{s_y}\) is the target range’s midpoint (Table~\ref{tab:style_attributes}).

    \item \textbf{STAR}: Represents the predicted style class (1--4 for readability, 1--5 for sentiment) using classifiers (e.g., DistilBERT for sentiment, FRE-to-level mapping for readability). \textbf{STAR}$_\Delta$ measures deviation from the target class:
    \begin{align*}
        \text{STAR}_\Delta = |\text{STAR}_{\hat{y}} - \text{STAR}_{s_y}|,
    \end{align*}
    where \(\text{STAR}_{\hat{y}}\) is the predicted class, and \(\text{STAR}_{s_y}\) is the target class.

    \item \textbf{RG-L}: Assesses content preservation via ROUGE-L, measuring the longest common subsequence between input and generated texts:
    \begin{align*}
        \text{RG-L} = \frac{\text{LCS}(X, \hat{Y})}{|\hat{Y}|},
    \end{align*}
    where \(\text{LCS}(X, \hat{Y})\) is the length of the longest common subsequence between input \(X\) and generated text \(\hat{Y}\), and \(|\hat{Y}|\) is the length of \(\hat{Y}\).

    \item \textbf{H-Re}: Evaluates overall transfer quality, equivalent to the reward functions for readability and sentiment in Section~\ref{sec:total reward}, but with the BERTScore coefficient set to 0:
    \begin{align*}
        \text{H-Re} = 0.5 \cdot \mathcal{R}_{\text{sent}} + 0.5 \cdot \mathcal{R}_{\text{lex}},
    \end{align*}
     where $\mathcal{R}_{\text{sent}}$ and $\mathcal{R}_{\text{lex}}$ are defined in Section~\ref{hierarchical rewards}.
\end{itemize}

\section{DataSet Overview}
\label{ap:dataset}

Table~\ref{tab:style_attributes} demonstrates the details of the readability and sentiment datasets, involving level scores, intensity attributes, and data distribution.

\begin{table*}[!h]
  \centering
  \renewcommand{\arraystretch}{1.2} %
  \begin{tabular}
  {cccccc}
    \hline
    \textbf{Style} & \textbf{Dataset} & \textbf{SCORES} & \textbf{ATTRIBUTES} & \textbf{Train} & \textbf{Test} \\
    \hline
    \multirow{4}{*}{Readability} & \multirow{4}{*}{CNN/DM} & $80 \leq \mathrm{FRE} \leq 100$ & Elementary School Students & 4,000 & \multirow{4}{*}{4,000} \\
                                 &                       & $60 \leq \mathrm{FRE} < 80$   & Middle School Students & 4,000 & \\
                                 &                       & $40 \leq \mathrm{FRE} < 60$   & High School Students & 4,000 & \\
                                 &                       & $0 \leq \mathrm{FRE} < 40$    & College Students & 4,000 & \\
    \hline
    \multirow{5}{*}{Sentiment}   & \multirow{5}{*}{Yelp}  & 5 stars                      & Very Positive Tone & 4,000 & \multirow{5}{*}{5,000} \\
                                 &                       & 4 stars                      & Positive Tone & 4,000 & \\
                                 &                       & 3 stars                      & Neutral Tone & 4,000 & \\
                                 &                       & 2 stars                      & Negative Tone & 4,000 & \\
                                 &                       & 1 star                       & Very Negative Tone & 4,000 & \\
    \hline
  \end{tabular}
  \caption{Attributes for Style, categorized by Readability and Sentiment. The number of training sets here refers to the number of constructed pseudo-parallel datasets.}
  \label{tab:style_attributes}
\end{table*}

\section{Representative words for stylistic intensities}
\label{sec:representative words}

As shown in Table~\ref{tab:feature words}, some representative words in the readability and sentiment corpora for stylistic intensities are listed, forming the vocabulary distribution for the lexicon-level reward.

\begin{table*}[h]
  \centering
  \begin{tabular}{ll}
    \hline
    \textbf{Scores} & \textbf{feature words} \\
    \hline
    \multicolumn{2}{c}{\textbf{Readability}} \\
    \hline
    elementary school & say, people, big, hurt, did, help, get, year, call, bad\\
    middle school & mention, just, believe, like, old, decide, share, know, right, important\\
    high school & state, according, express, additionally, currently, describe, recently, claim\\
    college & regarding, significant, individual, following, subsequently, notably, furthermore\\
    \hline
    \multicolumn{2}{c}{\textbf{Sentiment}} \\
    \hline
    very positive & outstanding, amazing, absolutely, incredible, truly, fantastic, gem, incredibly\\
    positive & solid, great, overall, definitely, enjoyable, satisfying, delightful, tasty, really\\
    neutral & bit, decent, quite, improvement, potential, overall, nice, consider, appreciate\\
    negative & disappointing, unfortunately, left, desired, frustrating, better, expected, lacking\\
    very negative & terrible, awful, bland, disaster, disappointment, avoid, worst, completely, barely\\
    \hline
  \end{tabular}
  \caption{Representative words in readability and sentiment corpora for stylistic intensities.}
  \label{tab:feature words}
\end{table*}

\section{Explanation of the decrease in RG-L}
\label{app:Rouge-L}
On the Readability dataset, the RG-L metric only exhibits a slight decrease. This phenomenon can be attributed to the core calculation logic of the metric: RG-L incorporates the Longest Common Subsequence (LCS) of the text before and after style transformation for analysis. The design of this metric focuses on measuring the degree of textual content preservation between the pre- and post-transformation texts. However, the requirement for such content preservation inevitably leads to changes in the lexical hierarchy, making the RG-L metric irrelevant to the core goal of "evaluating style transformation intensity". Thus, the effectiveness of style transformation methods cannot be comprehensively assessed solely based on this metric.

\subsection{Evidence from Tabular Case}
Table \ref{tab:RG-L} presents a specific empirical case that supports the aforementioned conclusion.

\begin{table*}[h]
  \centering
  \begin{adjustbox}{max width=\textwidth}
  \begin{tabular}{llll}
    \hline
    \textbf{Text Category} & \textbf{FRE} & \textbf{RG-L} & \textbf{Text Content} \\
    \hline
    Original Text & 66.10 & — & The scientist did careful tests to get correct results.\\
    Rewritten Text 1 & 33.07 & 66.67 & The scientist did careful experiments to obtain precise results.\\
    Rewritten Text 2 & 15.64 & 52.63 & The scientist carefully implemented the experiments to ensure precise results.\\
    \hline
  \end{tabular}
  \end{adjustbox}
  \caption{Example of Text Style Transformation with Different Intensity Levels}
  \label{tab:RG-L}

\end{table*}

\subsection{Analysis of Metric-Goal Adaptability}
The initial objective of this study was to rewrite the original text into a version that meets the "college-level and above reading standard", which corresponds to a FRE score range of 0–40. As shown in the table, both Rewritten Text 1 and Rewritten Text 2 satisfy this requirement.

Further analysis of style transformation intensity reveals that the FRE score of Rewritten Text 2 (15.64) is lower than that of Rewritten Text 1 (33.07). This numerical difference indicates that Rewritten Text 2 exhibits a more significant stylistic shift. Nevertheless, although the RG-L value of Rewritten Text 2 (52.63) is lower than that of Rewritten Text 1 (66.67), this difference in RG-L values does not diminish its superiority in style transformation effectiveness—this further confirms the conclusion that the RG-L metric cannot effectively reflect style transformation intensity.

\section{Hyperparameter Settings}
\label{app:hyperparameters}

Table~\ref{tab:hyperparameters} shows some settings for UTST with the SFT-then-PPO paradigm, including computing infrastructure, GPT-4o-mini settings, Deepspeed settings, SFT settings and PPO settings. 

\subsection{Setting Strategy of Hyperparameters for the Combined Reward Function}
\label{setting_strategy}
For the combined reward function, $\lambda_1$, $\lambda_2$, and $\lambda_3$ are configured via a fixed-weight strategy, excluding resource-intensive grid search. The weights adhere to an order-preserving principle, sorted by their significance: the sentence-level reward ($R_{\text{sent}}$) is assigned the highest weight, with the objective of prioritizing the alignment between the global style of generated text and the target intensity; the word-level reward ($R_{\text{lex}}$) follows, employed to enhance the matching between fine-grained lexical features and the target style; the consistency reward ($R_{\text{cons}}$) is assigned the lowest priority, as it is not the core focus of this study. This reward merely requires the preservation of core semantics during style transfer, and no ablation experiments were performed for this component.

\begin{table}[htbp]
\small
\centering
\begin{tabular}{ll}
\toprule
\textbf{Parameter} & \textbf{Value} \\
\midrule

\multicolumn{2}{c}{\textit{Computational Settings}} \\
GPU & NVIDIA A6000 48GB (2 units) \\
flan-t5-large & 783M parameters\\
Roberta-large & 355M parameters\\
DistilBERT base model & 67M parameters\\

\midrule

\multicolumn{2}{c}{\textit{GPT-4o-mini Settings}} \\
temperature & 0.7 \\

\midrule
\multicolumn{2}{c}{\textit{Deepspeed Settings}} \\
ZeRO Optimization Stage & 2 \\
Mixed Precision & bf16 \\
Gradient Accumulation Steps & 8 \\
Gradient Clipping & 1.0 \\
Communication Window Size & 512 \\

\midrule
\multicolumn{2}{c}{\textit{SFT Settings}} \\
Optimizer & AdamW \\
Optimizer Parameters & $\beta = (0.9, 0.999)$, $\epsilon = 10^{-8}$ \\
Learning Rate & $10^{-4}$ \\
Weight Decay & 0.0 \\
Dropout Rate & 0.1 \\
Batch Size & 16 \\
Epochs & 20 \\

\midrule
\multicolumn{2}{c}{\textit{PPO Settings}} \\
Optimizer & AdamW \\
Optimizer Parameters & $\beta = (0.9, 0.999)$, $\epsilon = 10^{-8}$ \\
Learning Rate & $10^{-5}$ \\
Weight Decay & $10^{-6}$ \\
Scheduler Parameters & $T_{\text{max}} = 10^4$, $\eta_{\text{min}} = 10^{-6}$ \\
Batch Size & 2 \\
Epochs & 10 \\
Total Steps & 100000 \\
Num Rollouts & 512 \\
Chunk Size & 4 \\
Initial KL Coefficient & 0.05 \\
Target KL Divergence & 6 \\
Horizon & 10000 \\
Discount Factor ($\gamma$) & 0.99 \\
GAE Lambda ($\lambda$) & 0.95 \\
Policy Clip Range & 0.2 \\
Value Clip Range & 0.2 \\
Value Function Coefficient & 1.0 \\
Reward Clip Range & 10 \\
Max New Tokens (Generation) & 256 \\
Sampling Temperature & 1.0 \\
Top-k Sampling & 50 \\
Top-p Sampling & 0.95 \\

\midrule
\multicolumn{2}{c}{\textit{General}} \\
Random Seed & 42 \\

\bottomrule
\end{tabular}
\caption{Hyperparameter settings}
\label{tab:hyperparameters}
\end{table}

\clearpage

\end{document}